\documentclass[11pt]{article}
\usepackage{coling2016}
\usepackage{times}
\usepackage{url}
\usepackage{latexsym}

\usepackage{framed}
\usepackage{scrextend}
\usepackage{booktabs}
\usepackage{amsmath}
\usepackage{paralist}
\usepackage{tikz}
\usetikzlibrary{positioning}

\graphicspath{{./img/}}
\DeclareGraphicsExtensions{.pdf,.jpeg,.png,.jpg}

\usepackage{color}

\title{Stance Classification in Rumours as a Sequential Task Exploiting the Tree Structure of Social Media Conversations}

\author{Arkaitz Zubiaga$^1$, Elena Kochkina$^1$, Maria Liakata$^1$, Rob Procter$^1$, Michal Lukasik$^2$ \\
   $^1$ University of Warwick, Coventry, UK \\
   $^2$ University of Sheffield, Sheffield, UK \\
   {\tt \{a.zubiaga,e.kochkina,m.liakata,rob.procter\}@warwick.ac.uk}\\
   {\tt m.lukasik@sheffield.ac.uk}}

\date{}

\begin{document}
\maketitle
\begin{abstract}
 Rumour stance classification, the task that determines if each tweet in a collection discussing a rumour is supporting, denying, questioning or simply commenting on the rumour, has been attracting substantial interest. Here we introduce a novel approach that makes use of the sequence of transitions observed in tree-structured conversation threads in Twitter. The conversation threads are formed by harvesting users' replies to one another, which results in a nested tree-like structure. Previous work addressing the stance classification task has treated each tweet as a separate unit. Here we analyse tweets by virtue of their position in a sequence and test two sequential classifiers, Linear-Chain CRF and Tree CRF, each of which makes different assumptions about the conversational structure. We experiment with eight Twitter datasets, collected during breaking news, and show that exploiting the sequential structure of Twitter conversations achieves significant improvements over the non-sequential methods. Our work is the first to model Twitter conversations as a tree structure in this manner, introducing a novel way of tackling NLP tasks on Twitter conversations.
\end{abstract}

\section{Introduction}

\blfootnote{
 \hspace{-0.65cm}  
 This work is licensed under a Creative Commons 
 Attribution 4.0 International Licence.
 Licence details:
 \url{http://creativecommons.org/licenses/by/4.0/}
}

Rumour stance classification is a task that is increasingly gaining popularity in its application to tweets. While Twitter is a generous source of reports of breaking news, outpacing even news outlets \cite{kwak2010twitter}, it also comes with the caveat that some of those reports are still rumours at the time of posting and so are yet to be verified and corroborated \cite{mendoza2010twitter,procter2013readinga,procter2013readingb}. The rumour stance classification task intends to assist in this verification process by determining the type of support expressed in different tweets discussing the same rumour \cite{qazvinian2011rumor}. Aggregation of the stance of multiple tweets discussing a rumour can then be of help to determine its likely veracity, enabling -- among other benefits -- the flagging of highly disputed rumours that are likely to be false.

Previous research on rumour stance classification for tweets has been limited to the tweet as the unit to be classified. However, such approaches ignore the additional context and knowledge that can be gained from the structure of Twitter interactions within conversational threads \cite{zubiaga2016analysing,procter2013readinga,tolmie2015microblog}. The latter are formed as Twitter users reply to one another's posts and ultimately users build on each others' stance towards the rumour, leading to a potential consensus. For example, a tweet may report a rumour (source tweet) and others may reply to it by further supporting it or providing counter-evidence. Our objective here is to mine the sequence of stance types encountered in conversational threads collected from Twitter. The ultimate goal could be to aggregate such views to help determine the veracity of a rumour, which we hypothesise our task could be helpful for.

In order to make use of the sequence of stance types, we analyse conversations arising from tweets posted by users who are replying to one another. These replies result in tree-structured conversations, often nested, where replies are triggered by a source tweet that initiated the conversation. We make the following contributions:

\begin{compactitem}
\item We hypothesise that making use of the sequential structure of conversational threads can improve stance classification in relation to a classifier that determines a tweet's stance from the tweet alone. To the best of our knowledge, the structure of Twitter conversations has not been studied before for classifying each of the underlying tweets and our work is the first to evaluate it for stance classification.

\item We introduce a novel way of analysing tweets by mining the context from conversational threads. To do this, we propose two different models for capturing the sequential structure of conversational threads, viewing them as a) separate linear branches and b) as a tree structure.

\item We evaluate the effectiveness of two flavours of Conditional Random Fields (CRF) in addressing stance classification on rumourous Twitter conversations. We compare the performance of these two CRF settings with other non-sequential baselines, including the non-sequential equivalent of CRF, a Maximum Entropy classifier. Our results show that while there is no significant difference when performance is measured based on micro-averaged F1 score (equivalent to accuracy and influenced by the majority class), sequential approaches do perform substantially better in terms of macro-averaged F1 score, proving that exploiting the conversational structure improves the classification performance.

\item We also show that the use of tree CRF leads to an improvement over the linear-chain CRF, suggesting that in stance classification for conversational threads it is important to consider the whole tree structure rather individual linear branches. Our results advocate the merit of further exploring the use of sequential approaches to exploit conversational structures mined from Twitter posts for a wider range of NLP tasks.
\end{compactitem}

\section{Related Work}

Following early work by Qazvinian et al. \shortcite{qazvinian2011rumor} introducing the task of rumour stance classification for tweets, interest in this problem has increased substantially. However, the line of research initiated by Qazvinian et al. \shortcite{qazvinian2011rumor} is significantly different to the one tackled in this paper. They perform 2-way classification of each tweet as \textit{supporting} or \textit{denying} a long-standing rumour, such as disputed beliefs that \textit{Barack Obama is reportedly Muslim}. The authors use tweets observed in the past to train a classifier, which is then applied to new tweets discussing the same rumour. In recent work, rule-based methods have been put forward as a way to improve on the performance of the Qazvinian et al. \shortcite{qazvinian2011rumor} baseline. This is the approach followed by Liu et al. \shortcite{liu2015realtime}, who introduced a simple rule-based method that looks for the presence of positive or negative words in a tweet. One draw back of such rule-based approaches is that they may not generalise to new, unseen rumours. Similarly, Hamidian and Diab \shortcite{hamidian2016rumor} have recently studied the extent to which a model trained from historical tweets can be used for classifying new tweets discussing the same rumour. While Zhao et al. \shortcite{zhao2015enquiring} did not study stance classification, they showed that tweets that trigger questioning responses from others are likely to report disputed rumours, which reinforces the motivation of our work of determining the stance of tweets to then deal with rumours.

Classification of stance towards a target on Twitter has been addressed in SemEval-2016 task 6 \cite{mohammad2016semeval}. Task A had to determine the stance of tweets towards five targets as `favor', `against' or `none'. Task B tested stance detection towards an unlabelled target, which required a weakly supervised or unsupervised approach. The dataset of this competition was not related to rumours or breaking news, it only considered a 3-way classification and did not provide any relations between tweets, which were treated as individual instances.

Our work presents different objectives in three aspects. First, we aim to classify the stance of tweets towards rumours that emerge while breaking news unfold; these rumours are unlikely to have been observed before, and hence rumours from previously observed events, which are likely to diverge, need to be leveraged for training. As far as we know, only Lukasik et al. \shortcite{lukasik2015classifying,lukasik2016using,lukasik2016hawkes} have tackled stance classification in the context of breaking news applied to new rumours. Lukasik et al. \shortcite{lukasik2015classifying,lukasik2016using} used Gaussian Processes to perform 3-way stance classification into supporting, denying or questioning, while comments where not considered as part of the task. Lukasik et al. \shortcite{lukasik2016hawkes} did include comments to perform 4-way stance classification; they used Hawkes Processes to exploit the temporal sequence of stances towards rumours to classify new tweets discussing rumours. Work by Zeng et al. \cite{zeng2016unconfirmed} has also performed stance classification for rumours around breaking news, but overlapping rumours were used for training and testing. 

Second, recent research has posited that a 4-way classification is needed to capture responses seen in the unfolding of breaking news \cite{procter2013readinga,zubiaga2016analysing}. Moving away from the 2-way classification above, which is somewhat limited for our purposes, we adopt this expanded scheme including tweets that are \textit{supporting}, \textit{denying}, \textit{querying} or \textit{commenting} rumours. This adds two more categories to the scheme used in early work, where tweets would only support or deny a rumour. Moreover, our approach takes into account the interaction between users on social media, whether it is about appealing for more information in order to corroborate a rumourous post (\textit{querying}) or to say something that does not contribute to the resolution of the rumour's veracity (\textit{commenting}). Finally, instead of dealing with tweets as single units in isolation, we exploit the conversational structure of Twitter replies, building a classifier that learns the dynamics of stance in tree-structured conversational threads. The closest work when it comes to exploiting conversational structure in tweets is that of Ritter et al. \shortcite{ritter2010unsupervised} who modelled linear sequences of replies in Twitter conversations with Hidden Markov Models for dialogue act tagging, but the structure of the tree as a whole was not exploited.

As far as we know, no work has leveraged the conversational structure of Twitter postings for stance classification, and hence its utility remains unexplored. A work that is related is that of Lukasik et al. \shortcite{lukasik2016hawkes}, who exploited the temporal sequence of tweets, although the conversational structure was ignored and each tweet was treated as a separate unit. In other domains where debates or conversations are involved, the sequence of responses has been exploited to make the most of the evolving discourse and perform an improved classification of each individual post after learning the structure and dynamics of the conversation as a whole. For instance, Qu and Liu \shortcite{qu2011finding} found Hidden Markov Models to be an effective approach to classify threads in on-line fora as successfully solving or not the question raised in the initial post. This was later further studied in a SemEval shared task, where each post in a forum thread had to also be classified as good, potential or bad \cite{marquez2015semeval}. FitzGerald et al. \shortcite{fitzgerald2011exploiting} used a linear-chain CRF to identify high-quality comments in threads responding to blog posts.

In a task that is related to stance classification, researchers have also studied the identification of agreement and disagreement in on-line conversations. To classify agreement between question-answer (Q-A) message pairs in fora, Abbott et al. \shortcite{abbott2011can} used Naive Bayes as the classifier, and Rosenthal and McKeown \shortcite{rosenthal2015couldn} used a logistic regression classifier. However, in both cases only pairs of messages were considered, and the entire sequence of responses in the tree was not used. CRF has also been used to detect agreement and disagreement between speakers in broadcast debates \cite{wang2011detection}, which our task differs from in that it solely focuses on text. It is also worthwhile to emphasise that stance classification is different to agreement/disagreement detection, given that in stance classification one has to determine the orientation of a user towards a rumour. Instead, in agreement/disagreement detection, one has to determine if a pair of posts share the same view. In stance classification, one might agree with another user who is denying a rumour, and hence they are denying the rumour as well, irrespective of the pairwise agreement.
To the best of our knowledge Twitter conversational thread structure has not been explored in the stance classification problem.

\section{Stance classification using the conversational structure of Twitter threads}

The rumour stance classification task consists in determining the type of support that each individual post expresses towards the disputed veracity of a rumour. The task is especially interesting in the context of Twitter, where unverified reports about breaking news are continually being posted and discussed as they unfold. This problem was originally tackled as a 2-way classification task, where each tweet was classified as supporting or denying a rumour. However, recent research \cite{procter2013readinga} found this categorisation to be insufficient to encompass all the different kinds of reactions to rumours, and a broader, 4-way classification task has been suggested instead. The argument behind this is that users in social media will not necessarily express a clear inclination towards supporting or denying a rumour, but can also be skeptical by posing questions about it or can make comments about the rumour that are unrelated to its disputed veracity. The four categories in the extended classification scheme thus include \textit{supporting}, \textit{denying}, \textit{querying} and \textit{commenting}. In this work we set out the rumour stance classification that adopts this broader scheme. We define the rumour stance classification task as follows: we assume we have a set $D$ of rumours $R_i$, each of which is composed of a collection of rumourous conversational threads. For simplicity here we refer to a rumour as the aggregate of its Twitter conversational threads, which is ultimately a collection of tweets. Each rumour has a variably sized set of tweets $t_i$ discussing it so that $R_i = \{t_1, ..., t_{|R_i|}\}$; the task consists in determining the stance of each of the tweets $t_j$ pertaining to a new, unseen rumour $R_i$ as one of $Y = \{supporting, denying, querying, commenting\}$.

Moreover, within this task we propose leveraging conversation structure as one of the main features that characterise social media \cite{tolmie2015microblog}. So the task becomes one of classifying each tweet in a conversational thread, in the context of the thread. Twitter conversations consist of replies to each other, together forming a tree structure, as shown in the example in Figure \ref{fig:example}. Replies can be nested in each other, so that the depth of the tree can vary. Hence, in the stance classification task applied to Twitter conversations we have rumours containing a variably sized set of conversations $R_i = \{C_1, ..., C_{|R_i|}\}$. Each of these conversations, $C_j$, has a varying number of tweets in it. By definition, a conversation has a source tweet (the root of the tree), $t_{j,1}$, that initiates it. The source tweet $t_{j,1}$ can receive replies by a varying number $k$ of tweets $Replies_{t_{j,1}} = \{t_{j,1,1}, ..., t_{j,1,k}\}$, each of which can in turn receive replies by a varying number $k$ of tweets, e.g., $Replies_{t_{j,1,1}} = \{t_{j,1,1,1}, ..., t_{j,1,1,k}\}$. 
Thus, we encode the tweet index as a sequence of ids of consecutive children of a preceding node, while traversing the conversation structure.

\section{Dataset}

We use the PHEME rumour dataset associated with eight events corresponding to breaking news stories \cite{zubiaga2016analysing}, which provide tweet-level annotations for stance\footnote{While the dataset includes data for nine events, here we use the eight events whose tweets are in English, excluding the ninth with tweets in German.}. Tweets in this dataset include tree-structured conversations, which are initiated by a tweet about a rumour (source tweet) and nested replies that further discuss the rumour circulated by the source tweet (replying tweets). Details on how the annotation was conducted through crowdsourcing can be found in Zubiaga et al. \shortcite{zubiaga2015crowdsourcing}.

The annotation scheme employed by the authors differs slightly from the one we need for our purposes, so we adapt it to our needs as follows. The source tweet of a conversation is originally annotated as \emph{supporting} or \emph{denying}, and each subsequent tweet is annotated as \emph{agreed}, \emph{disagreed}, \emph{appeal for more information} (\emph{querying}) or \emph{commenting} as a pairwise annotation with respect to the source tweet. Instead, the labels needed for our task are \textit{supporting}, \textit{denying}, \textit{querying} and \textit{commenting}. To convert the labels, we keep the labels as \textit{supporting} or \textit{denying} in the case of source tweets. For the reply tweets, we keep their label as is for the tweets that are \textit{querying} or \textit{commenting}. To convert those tweets that agree or disagree into \textit{supporting} or \textit{denying}, we apply the following set of rules: (1) if a tweet agrees with a supporting source tweet, we label it \textit{supporting}, (2) if a tweet agrees with a denying source tweet, we label it \textit{denying}, (3) if a tweet disagrees with a supporting source tweet, we label it \textit{denying} and (4) if a tweet disagrees with a denying tweet, we label it \textit{supporting}. The latter enables us to infer stance with respect to the overarching rumour rather than refer to agreement with respect to the source. The resulting dataset includes 4,519 tweets, and the transformations of annotations described above only affect 24 tweets (0.53\%), i.e., those where the source tweet denies a rumour, which is rare. The example in Figure \ref{fig:example} shows a rumour thread taken from the dataset along with our inferred annotations, as well as how we establish the depth value of each tweet in the thread.

\begin{figure*}
 \begin{framed}
  \textit{[depth=0]} \noindent \textbf{u1:} These are not timid colours; soldiers back guarding Tomb of Unknown Soldier after today's shooting \#StandforCanada --PICTURE-- \textbf{[support]}
  \begin{addmargin}[2em]{0pt}
   \textit{[depth=1]} \textbf{u2:} @u1 Apparently a hoax. Best to take Tweet down. \textbf{[deny]}
  \end{addmargin}
  \begin{addmargin}[2em]{0pt}
   \textit{[depth=1]} \textbf{u3:} @u1 This photo was taken this morning, before the shooting. \textbf{[deny]}
  \end{addmargin}
  \begin{addmargin}[2em]{0pt}
   \textit{[depth=1]} \textbf{u4:} @u1 I don't believe there are soldiers guarding this area right now. \textbf{[deny]}
  \end{addmargin}
  \begin{addmargin}[4em]{0pt}
   \textit{[depth=2]} \textbf{u5:} @u4 wondered as well. I've reached out to someone who would know just to confirm that. Hopefully get response soon. \textbf{[comment]}
  \end{addmargin}
  \begin{addmargin}[6em]{0pt}
   \textit{[depth=3]} \textbf{u4:} @u5 ok, thanks. \textbf{[comment]}
  \end{addmargin}
 \end{framed}
 \caption{Example of a tree-structured thread discussing the veracity of a rumour, where the label associated with each tweet is the target of the rumour stance classification task.}
 \label{fig:example}
\end{figure*}

One notable characteristic of the dataset is that the distribution of categories is skewed towards \textit{commenting} tweets, and that this imbalance varies slightly across the eight events in the dataset (see Table \ref{tab:dataset-stats}). Given that we consider each event as a separate fold that is left out for testing, this varying imbalance makes the task more realistic and challenging. The striking imbalance towards \textit{commenting} tweets is also indicative of the increased difficulty with respect to previous work on stance classification. Most of which performed binary classification of tweets as either supporting or denying, which as shown in our experiments only account for less than 28\% of the tweets.

\begin{table}[htb]
 \centering
 \begin{tabular}{ l c c c c c }
  \toprule
  Event & Supporting & Denying & Querying & Commenting & Total \\
  \midrule
  charliehebdo & 239 (22.0\%) & 58 (5.0\%) & 53 (4.0\%) & 721 (67.0\%) & 1,071 \\
  ebola-essien & 6 (17.0\%) & 6 (17.0\%) & 1 (2.0\%) & 21 (61.0\%) & 34 \\
  ferguson & 176 (16.0\%) & 91 (8.0\%) & 99 (9.0\%) & 718 (66.0\%) & 1,084 \\
  germanwings-crash & 69 (24.0\%) & 11 (3.0\%) & 28 (9.0\%) & 173 (61.0\%) & 281 \\
  ottawashooting & 161 (20.0\%) & 76 (9.0\%) & 63 (8.0\%) & 477 (61.0\%) & 777 \\
  prince-toronto & 21 (20.0\%) & 7 (6.0\%) & 11 (10.0\%) & 64 (62.0\%) & 103 \\
  putinmissing & 18 (29.0\%) & 6 (9.0\%) & 5 (8.0\%) & 33 (53.0\%) & 62 \\
  sydneysiege & 220 (19.0\%) & 89 (8.0\%) & 98 (8.0\%) & 700 (63.0\%) & 1,107 \\
  \midrule
  Total & 910 (20.1\%) & 344 (7.6\%) & 358 (7.9\%) & 2,907 (64.3\%) & 4,519 \\
  \bottomrule
 \end{tabular}
 \caption{Distribution of categories for the eight events in the dataset.}
 \label{tab:dataset-stats}
\end{table}

\section{Experiment Design}

In this section we describe the classifiers, features and evaluation measures we used in our experiments.

\subsection{Classifiers}


\textbf{Conditional Random Fields (CRF).} We use CRF as a structured classifier to model sequences observed in Twitter conversations. With CRF, we can model the conversation as a graph that will be treated as a sequence of stances, which also enables us to assess the utility of harnessing the conversational structure for stance classification. In contrast to traditionally used classifiers for this task, which choose a label for each input unit (e.g. a tweet), CRF also consider the neighbours of each unit, learning the probabilities of transitions of label pairs to be followed by each other. The input for CRF is a graph $G = (V, E)$, where in our case each of the vertices $V$ is a tweet, and the edges $E$ are relations of tweets replying to each other. Hence, having a data sequence $X$ as input, CRF outputs a sequence of labels $Y$ \cite{lafferty2001conditional}, where the output of each element $y_i$ will not only depend on its features, but also on the probabilities of other labels surrounding it. The generalisable conditional distribution of CRF is shown in Equation \ref{eq:crf} \cite{sutton2011introduction}.

\begin{equation}
 p(y|x) = \frac{1}{Z(x)} \prod_{a = 1}^{A} \Psi_a (y_a, x_a)
 \label{eq:crf}
\end{equation}

where Z(x) is the normalisation constant, and $\Psi_a$ is the set of factors in the graph $G$.

\begin{figure*}[ht]
 \centering
 \includegraphics[width=0.7\textwidth]{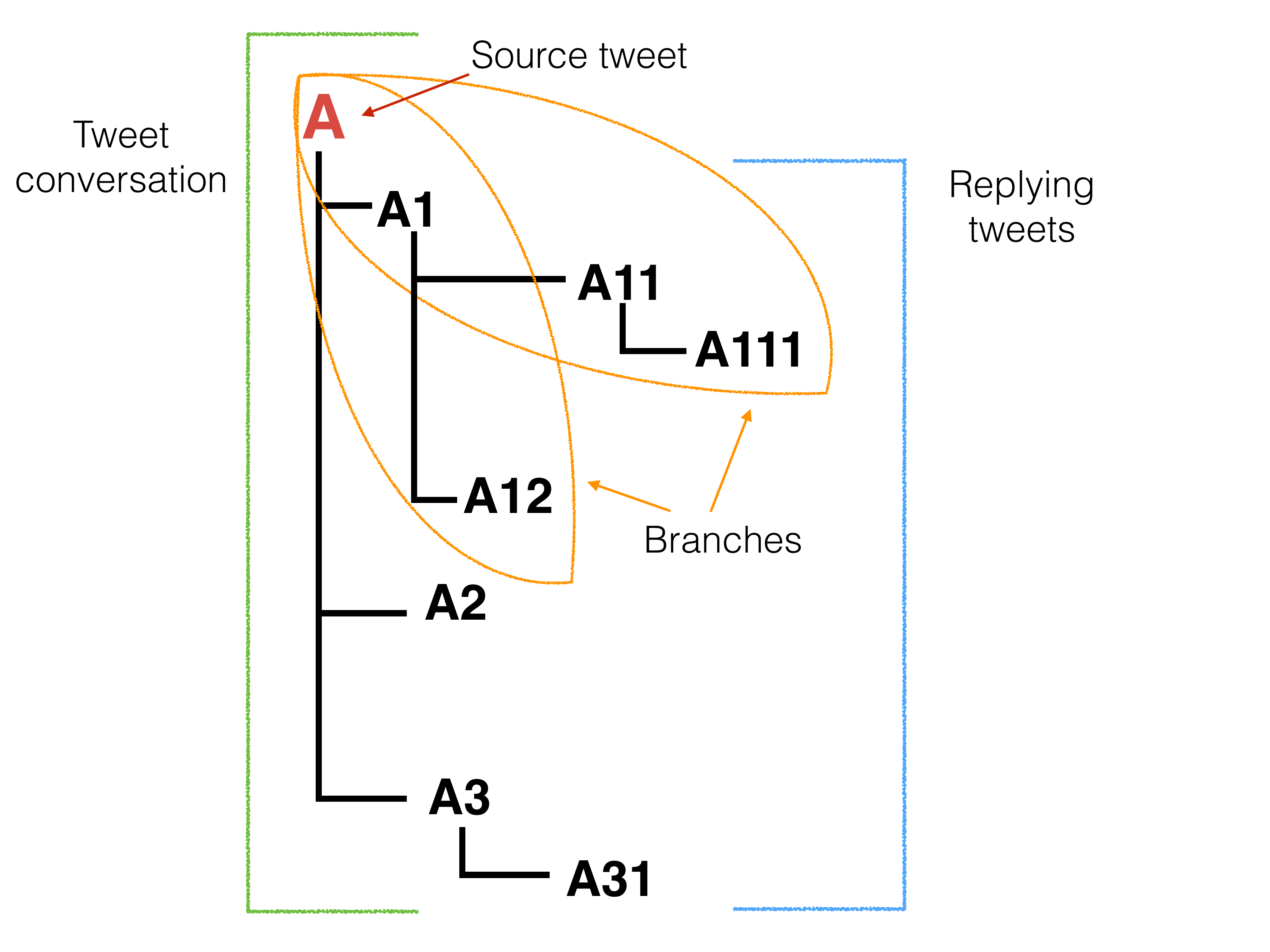}
 \caption{Example of a tree-structured conversation, with two overlapping branches highlighted.}
 \label{fig:tree-and-branches}
\end{figure*}

We use CRFs in two different settings.\footnote{We use the PyStruct to implement both variants of CRF \cite{muller2014pystruct}.} First, we use a linear-chain CRFs (Linear CRF) to model each branch as a sequence to be input to the classifier. We also use Tree-Structured CRFs (Tree CRF) or General CRFs to model the whole, tree-structured conversation as the sequence input to the classifier. So in the first case the sequence unit is a branch and our input is a collection of branches and in the second case our sequence unit is an entire conversation, and our input is a collection of trees. An example of the distinction of dealing with branches or trees is shown in Figure \ref{fig:tree-and-branches}. With this distinction we also want to experiment whether it is worthwhile building the whole tree as a more complex graph, given that users replying in one branch might not have necessarily seen and be conditioned by tweets in other branches. However, we believe that the tendency of types of replies observed in a branch might also be indicative of the distribution of types of replies in other branches, and hence useful to boost the performance of the classifier when using the tree as a whole. An important caveat of modelling a tree in branches is also that there is a need to repeat parts of the tree across branches, e.g., the source tweet will repeatedly occur as the first tweet in every branch extracted from a tree.\footnote{Despite this also leading to having tweets repeated across branches in the test set and hence producing an output repeatedly for the same tweet with Linear CRF, this output is consistent and there is no need to aggregate different outputs.}

\textbf{Maximum Entropy classifier (MaxEnt).} As the non-sequential equivalent of CRF, we use a Maximum Entropy (or logistic regression) classifier, which is also a conditional classifier but which operate at the tweet level, ignoring the conversational strucsture. This enables us to compare directly the extent to which treating conversations as sequences instead of having each tweet as a separate unit can boost the performance of the classifier.

\textbf{Additional baselines.} We also compare four more non-sequential classifiers\footnote{We use their implementation in the scikit-learn Python package}: Naive Bayes (NB), Support Vector Machines (SVM), Random Forests (RF), and Majority (i.e., a dummy classifier always labelling the most frequent class).

We experiment in an 8-fold cross-validation setting. Seven events are used for training and the remainder event is used for testing. With this, we simulate a realistic scenario where we need to use knowledge from past events to train a model that will be used to classify tweets in new events. For evaluation purposes, we aggregate the output of all eight runs as the micro-averaged evaluation across runs.

\subsection{Features}

We use four types of features to represent the tweets. Note that all of them are local features extracted from the tweet itself and independent of the rest of the conversation, hence enabling us to focus our comparison on how using the sequential structure can impact on the results.

\textbf{Feature type \#1: Lexicon.}
\begin{compactitem}
 \item \textit{Word Embeddings:} a vector with 300 dimensions averaging vector representations of the words in the tweet using Word2Vec \cite{mikolov2013distributed}. The Word2Vec model for each of the eight folds is trained from the collection of tweets pertaining to the seven events in the training set, so that the event (and the vocabulary) in the test set is unknown.
 \item \textit{Part of speech (POS) tags:} a vector where each feature represents the number of occurrences of a type of POS tag in the tweet. The vector is then composed of the numbers of occurrences of different POS tags in the tweet, parsed using Twitie \cite{bontcheva2013twitie}.
 \item \textit{Use of negation:} binary feature determining if a tweet has a negation word or not. We use a list of negation words, including: not, no, nobody, nothing, none, never, neither, nor, nowhere, hardly, scarcely, barely, don't, isn't, wasn't, shouldn't, wouldn't, couldn't, doesn't.
 \item \textit{Use of swear words:} binary feature determining if `bad' words are present in a tweet. We use a list of 458 bad words\footnote{\url{http://urbanoalvarez.es/blog/2008/04/04/bad-words-list/}}.
\end{compactitem}

\textbf{Feature type \#2: Content formatting.}
\begin{compactitem}
 \item \textit{Tweet length:} the length of the tweet in number of characters.
 \item \textit{Capital ratio:} the ratio of capital letters among all alphabetic characters in the tweet.
 \item \textit{Word count:} the number of words in the tweet, counted as the number of space-separated tokens.
\end{compactitem}

\textbf{Feature type \#3: Punctuation.}
\begin{compactitem}
 \item \textit{Use of question mark:} binary feature for the presence or not of question marks in the tweet.
 \item \textit{Use of exclamation mark:} binary feature for the presence or not of exclamation marks in the tweet.
 \item \textit{Use of period:} binary feature for the presence or not of periods in the tweet.
\end{compactitem}

\textbf{Feature type \#4: Tweet formatting.}
\begin{compactitem}
 \item \textit{Attachment of URL:} binary feature, capturing the use or not of URLs in the tweet.
 \item \textit{Attachment of picture:} binary feature that determines if the tweet has a picture attached.
 \item \textit{Is source tweet:} binary feature determining if the tweet is a source tweet or is instead replying to someone else. Note that this feature can also be extracted from the tweet itself, checking if the tweet content begins with a Twitter user handle or not; there is no need to make use of the conversational structure to extract this feature.
\end{compactitem}

\subsection{Evaluation Measures}

Given that the classes are clearly imbalanced in our case, evaluation solely based on accuracy can arguably suffice to capture competitive performance beyond the majority class. To account for the imbalance of the categories, we use both micro-averaged and macro-averaged F1 scores. Note that the micro-averaged F1 score is equivalent to the accuracy measure, while the macro-averaged F1 score complements it by measuring performance assigning the same weight to each category.

\section{Results}

Table \ref{tab:results} shows the results comparing performance of the different classifiers, both in terms of micro- and macro-F1 scores, and F1 scores by class. Due to the fact that the dataset is clearly imbalanced with a skew towards \textit{commenting} tweets, we observe that even the majority classifier performs very well in terms of micro-averaged F1 score. In fact, the majority classifier is only slightly outperformed by other classifiers if we look at this evaluation measure. This is why we argue for an evaluation based on macro-averaged F1 score, which accounts for the ability of classifiers to produce an output that better fits to the distribution of classes. Interestingly, we observe that the conditional classifiers (i.e., MaxEnt, Linear CRF and Tree CRF) perform substantially better than the rest in terms of macro-averaged F1 score, which are the only ones to achieve a score of at least 0.4. Comparison of macro-averaged F1 scores of these three classifiers shows that the Tree CRF slightly outperforms the Linear CRF, while both perform significantly better than the non-sequential Maximum Entropy classifier. These results therefore do suggest that exploiting the sequential structure of conversations can lead to improvements on stance classification in rumourous Twitter conversations using the same set of local features.

When we look at the performance by class, we can observe that classifiers performing well only in terms of micro-averaged F1 have the tendency to perform well for the majority class (comments). Interestingly, CRF classifiers using conversational structure show remarkable improvements for other classes, especially supporting and querying tweets, where Tree CRF performs the best. However, all classifiers struggle to classify denials, with performance scores comparable to the other categories. We believe that one of the main reasons for this is that denials are one of the minority classes in the dataset. While querying tweets are also rare, some of the features like question marks are highly indicative of a tweet being a query, and hence they are easier to classify. Denials may in turn have significant commonalities with comments, given that the latter may also use negating words which may seem like denials. As shown in the confusion matrix for the Tree CRF in Table \ref{tab:confusion}, the majority class \textit{commenting} is being chosen in as many as 75.8\% of the cases by the classifier for those tweets that are actually denials. Collection of additional denying tweets may be of help to improve performance in this class.

\begin{table}[htb]
 \centering
 \begin{tabular}{l || r r || r r r r }
  \toprule
  Classifier & Micro-F1 & Macro-F1 & S & D & Q & C \\
  \midrule
  Majority & 0.643 & 0.196 & 0.000 & 0.000 & 0.000 & 0.783 \\
  SVM & \textbf{0.676} & 0.292 & 0.372 & 0.000 & 0.000 & \textbf{0.796} \\
  Random Forest & 0.666 & 0.357 & 0.360 & 0.022 & 0.260 & 0.787 \\
  \midrule
  Naive Bayes & 0.175 & 0.203 & 0.435 & \textbf{0.147} & 0.169 & 0.060 \\
  MaxEnt & 0.666 & 0.400 & 0.352 & 0.062 & 0.396 & 0.789 \\
  Linear CRF & 0.646 & 0.433 & 0.454 & 0.105 & 0.405 & 0.767 \\
  Tree CRF & 0.655 & \textbf{0.440} & \textbf{0.462} & 0.088 & \textbf{0.435} & 0.773 \\
  \bottomrule
 \end{tabular}
 \caption{Micro- and Macro-F1 performance results, and F1 scores by class (S: supporting, D: denying, Q: querying, C: commenting)}
 \label{tab:results}
\end{table}

\begin{table}[htb]
 \centering
 \begin{tabular}{r || r r r r}
  \toprule
  & S & D & Q & C \\
  \midrule
  S & \textbf{366 (40.4\%)} & 32 (3.5\%) & 22 (2.4\%) & 487 (53.7\%) \\
  D & 38 (11.1\%) & \textbf{22 (6.4\%)} & 23 (6.7\%) & 260 (75.8\%) \\
  Q & 11 (3.1\%) & 10 (2.8\%) & \textbf{149 (41.6\%)} & 188 (52.5\%) \\
  C & 261 (9.0\%) & 91 (3.1\%) & 133 (4.6\%) & \textbf{2,421 (83.3\%)} \\
  \bottomrule
 \end{tabular}
 \caption{Confusion matrix for Tree CRF (S: supporting, D: denying, Q: querying, C: commenting).}
 \label{tab:confusion}
\end{table}

For comparison with the state-of-the-art stance classification approach by Lukasik et al. \shortcite{lukasik2016hawkes}, we present results broken down by event in Table \ref{tab:results-by-event}, both for their approach based on Hawkes Processes as well as our Tree CRF approach. Note that Lukasik et al. \shortcite{lukasik2016hawkes} only tested their approach on four of the events, and therefore performance scores for the rest of the events are not shown. As can be observed from the four events for which we have comparable results, the Hawkes Process performs better in terms of micro-F1, and therefore accurately classifying more instances. However, the Tree CRF performs substantially better in terms of macro-F1, which shows Tree CRF's ability to better estimate the distribution of labels in what is a highly imbalanced task and hence favouring the use of conversational structure in the classification process. We deem this a strong factor in this case as even a simple majority classifier achieves high micro-F1 scores, and the challenge lies in boosting macro-F1 scores to better balance the classification.

\begin{table}[htb]
 \centering
 \begin{tabular}{l || r r || r r }
  \toprule
  & \multicolumn{2}{c ||}{Tree CRF} & \multicolumn{2}{c}{HP \cite{lukasik2016hawkes}} \\
  \midrule
  Event & Micro-F1 & Macro-F1 & Micro-F1 & Macro-F1 \\
  \midrule
  ottawashooting & 0.629 & \textbf{0.457} & \textbf{0.678} & 0.323 \\
  ferguson & 0.559 & \textbf{0.390} & \textbf{0.684} & 0.260 \\
  charliehebdo & 0.686 & \textbf{0.427} & \textbf{0.729} & 0.326 \\
  sydneysiege & 0.677 & \textbf{0.495} & \textbf{0.686} & 0.325 \\
  germanwings-crash & 0.694 & 0.523 & -- & -- \\
  putinmissing & 0.660 & 0.446 & -- & -- \\
  prince-toronto & 0.670 & 0.518 & -- & -- \\
  ebola-essien & 0.629 & 0.384 & -- & -- \\
  \bottomrule
 \end{tabular}
 \caption{Micro- and Macro-F1 performance results broken down by event, along with a comparison with the results obtained by Lukasik et al. \shortcite{lukasik2016hawkes}'s state-of-the-art approach based on Hawkes Processes, where available.}
 \label{tab:results-by-event}
\end{table}

\begin{figure*}
  \centering
  \includegraphics[width=1\textwidth]{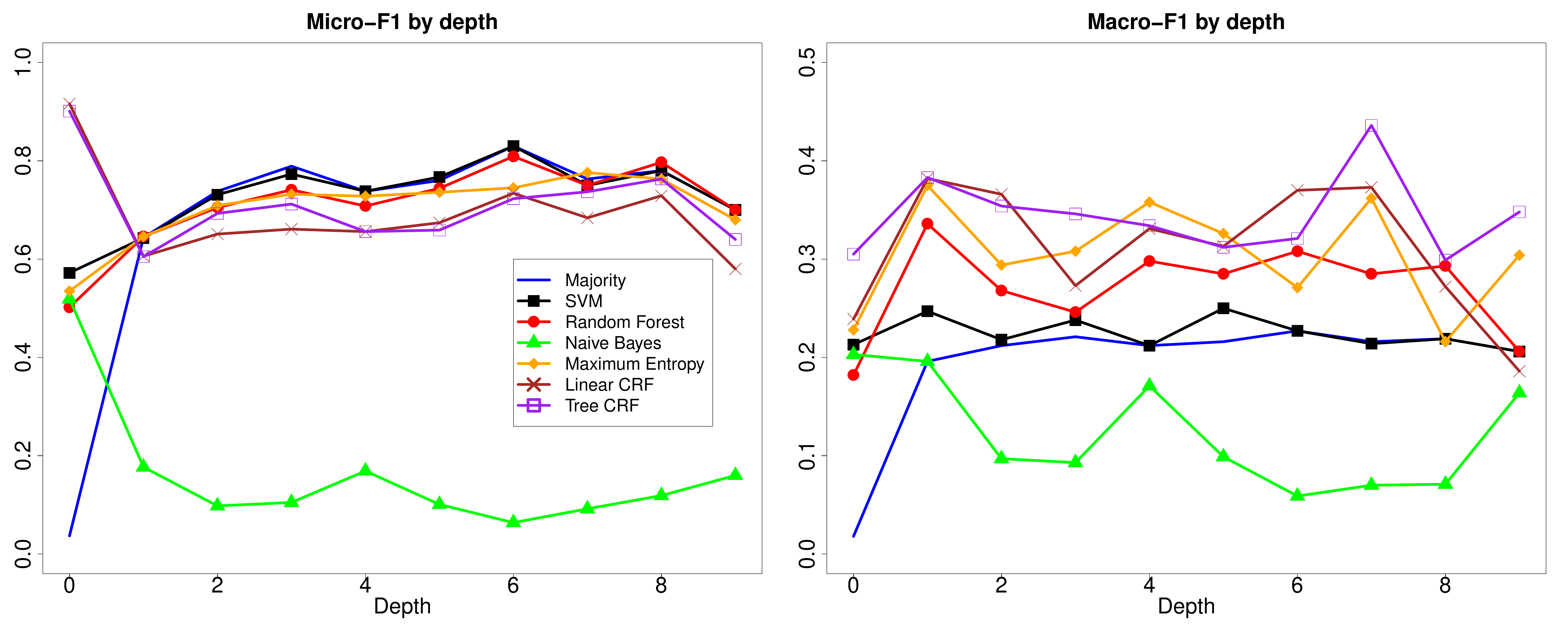}
  \caption{Micro- and macro-F1 scores by depth of tweet.}
  \label{fig:results-by-depth}
\end{figure*}

To better understand the effect of exploiting sequential structure, we break down performance scores by the depth of tweets. By this we want to see if the sequential classifiers are consistently performing well across tweets of different depth within conversations. Figure \ref{fig:results-by-depth} shows these results for tweets from depth 0 (source tweet) to depth 9. Further depths are omitted due to the small number of instances available. When we look at micro-averaged scores, we do not see a big performance difference across classifiers, except for the CRF classifiers performing better for source tweets; this is due to the fact that most of the source tweets tend to support a rumour, and hence sequential classifiers can learn this.

What is more interesting is to look again at the macro-averaged scores, where we see that the sequential approaches, especially the Tree CRF, consistently performs well for different levels of depth. More specifically, Tree CRF performs best in 7 out of 10 levels of depth analysed, with Linear CRF being better once (depth = 2) and Maximum Entropy being better twice (depth = 4 and 5).

\section{Conclusions}

We have introduced a novel way of tackling the rumour stance classification task, where a classifier has to determine if each tweet is supporting, denying, querying or commenting on a rumour's truth value. We mine the sequential structure of Twitter conversations in the form of users' replies to one another, extending existing approaches that treat each tweet as a separate unit. We have used two different sequential classifiers: a linear-chain CRF modelling tree-structured conversations broken down into branches, and a tree CRF modelling them as a graph that includes the whole tree. These classifiers have been compared with the non-sequential equivalent Maximum Entropy classifier, as well as other baseline classifiers, on eight Twitter datasets associated with breaking news.

While previous work had looked at the tweet as a single unit, we have shown that exploiting the discursive characteristics of interactions on Twitter, by considering probabilities of transitions within tree-structured conversational threads, can lead to significant improvements. Not only do we see that the linear sequence in a branch can be useful for the classifier to learn transitions, but also that having the whole picture of the tree showing the overall tendency of a conversation can further boost the performance of the classifier. Our results suggest that a tree CRF classifier outperforms all non-sequential classifiers, proving the utility of mining the conversational structure for stance classification, even when only local features are used.

To the best of our knowledge, this is the first attempt at aggregating the conversational structure of Twitter threads to produce classifications at the tweet level. Besides the utility of mining sequences from conversations for stance classification, we believe that our results will, in turn, encourage the study of sequential classifiers applied to other NLP tasks where the output for each tweet can benefit from the structure of the entire conversation, e.g., sentiment analysis and language identification.

Our plans for future work include testing additional sequential classifiers (e.g. LSTM). Moreover, while we have only tested local features for the purposes of making the experiments comparable, we also plan to test contextual features. This may also alleviate the effect of the class imbalance, producing results that are more satisfactory for minority classes, especially denials. Our approach assumes that rumours have been already identified or input by a human. An ambitious avenue for future work includes developing a rumour detection system whose output would be fed to the stance classification system.

\section*{Acknowledgments}

This work has been supported by the PHEME FP7 project (grant No. 611233). This research utilised Queen Mary's MidPlus computational facilities, supported by QMUL Research-IT and funded by EPSRC grant EP/K000128/1.

\bibliographystyle{acl}
\bibliography{sdqc}

\end{document}